%% file: main.tex
\pdfoutput=1
\documentclass[10pt,twocolumn]{article}

\usepackage[utf8]{inputenc}
\usepackage[T1]{fontenc}
\usepackage{lmodern}
\usepackage[a4paper,margin=0.75in]{geometry}
\usepackage{graphicx}
\usepackage{booktabs}
\usepackage{amsmath,amssymb}
\usepackage[table]{xcolor}
\usepackage[numbers,sort&compress]{natbib}
\usepackage[colorlinks=true,linkcolor=blue,citecolor=blue,urlcolor=blue]{hyperref}

\title{Reasoning About Traversability: Language-Guided Off-Road 3D Trajectory Planning}

\author{%
Byounggun Park \quad\quad Soonmin Hwang\thanks{Corresponding author: soonminh@hanyang.ac.kr}\\
Department of Automotive Engineering, Hanyang University, Seoul, Republic of Korea\\
\texttt{okharry1@hanyang.ac.kr, soonminh@hanyang.ac.kr}
}

\date{}

\begin{document}
\maketitle

\begin{abstract}
\input{sections/0_abstract.tex}
\end{abstract}

\noindent\textbf{Keywords:} Off-Road Autonomous Driving, Vision--Language Models, Trajectory Prediction, Language-Guided Planning, Terrain-Aware Navigation, Multimodal Data Refinement

\section{Introduction}
\input{sections/1_Introduction.tex}

\section{Related Work}
\input{sections/2_related_work.tex}

\section{Method}
\input{sections/3_method.tex}

\section{Experiments}
\input{sections/4_experiments.tex}

\section{Conclusion}
\input{sections/5_conclusion.tex}

\bibliographystyle{plainnat}
\bibliography{reference}

\end{document}

%% file: sections/0_abstract.tex
While Vision-Language Models (VLMs) enable high-level semantic reasoning for end-to-end autonomous driving, particularly in unstructured environments, existing off-road datasets suffer from language annotations that are weakly aligned with vehicle actions and terrain geometry. To address this misalignment, we propose a language refinement framework that restructures annotations into action-aligned pairs, enabling a VLM to generate refined scene descriptions and 3D future trajectories directly from a single image. To further encourage terrain-aware planning, we introduce a preference optimization strategy that constructs geometry-aware hard negatives and explicitly penalizes trajectories inconsistent with local elevation profiles. Furthermore, we propose off-road-specific metrics to quantify traversability compliance and elevation consistency, addressing the limitations of conventional on-road evaluation. Experiments on the ORAD-3D benchmark demonstrate that our approach reduces average trajectory error from 1.01m to 0.97m, improves traversability compliance from 0.621 to 0.644, and decreases elevation inconsistency from 0.428 to 0.322, highlighting the efficacy of action-aligned supervision and terrain-aware optimization for robust off-road driving.

%% file: sections/1_Introduction.tex
\begin{figure*}[t]
  \centering
  \includegraphics[width=\linewidth]{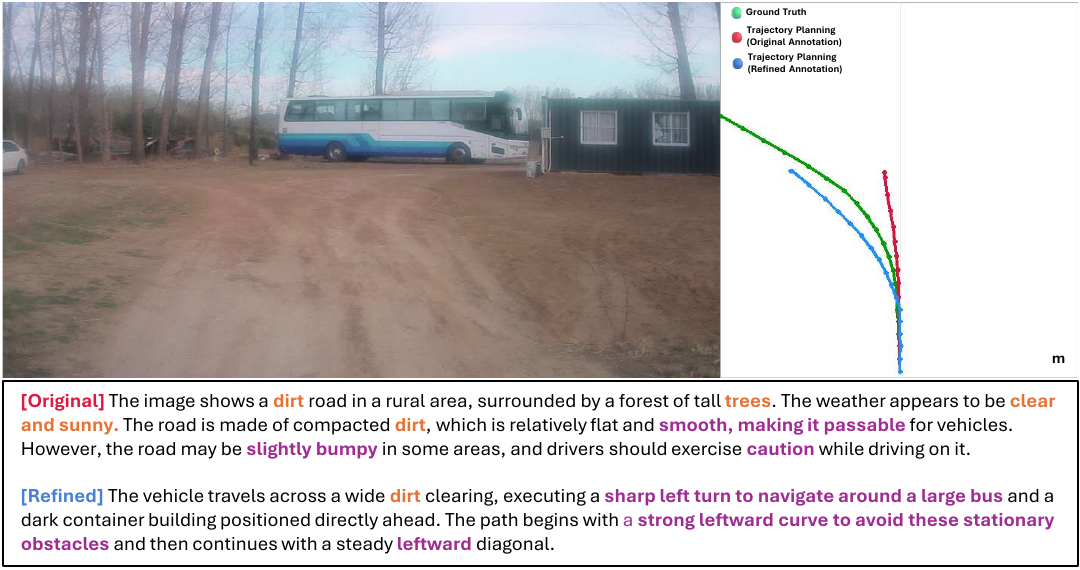}
  \caption{Qualitative comparison between SFT and Refined SFT. Refined language
  annotations are better aligned with trajectory geometry, resulting in more
  consistent motion and fewer failures.}
  \label{fig:sft_refine}
\end{figure*}

As driving environments become more diverse and unstructured, purely data-driven end-to-end models face challenges in reasoning about complex scenes, long-horizon decision-making, and rare or unseen situations~\cite{Chen_Wu_Chitta_Jaeger_Geiger_Li_2023}. To address these challenges, the research community is increasingly adopting Vision-Language Models (VLMs)~\cite{Li_Li_Yang_Yang_Chi_Yang_2025}, which combine strong visual understanding with the high-level semantic and reasoning capabilities inherited from large language models~\cite{Chen_Huang_Ma_Fang_Shi_Li_2025}. By leveraging language as an intermediate reasoning space, VLM-based driving systems have shown promising results in interpreting complex scenes, following high-level instructions, and improving robustness in corner cases~\cite{Tian_Gu_Li_Liu_Hu_Wang_Zhan_Jia_Lang_Zhao_2024}.

While VLM-based approaches have achieved substantial progress on established urban driving datasets such as nuScenes~\cite{nuscenes} and OpenScenes~\cite{openscene2023}, their direct applicability to off-road environments remains limited. Success on these urban datasets largely relies on strong environmental priors—such as lane markings, traffic signals, and well-defined road boundaries—that simplify both perception and planning. In contrast, off-road driving environments lack such explicit structures. As illustrated in Fig.~\ref{fig:sft_refine}, vehicles must instead reason about traversability, terrain geometry, obstacles, and surface conditions under severe visual ambiguity. Consequently, decision-making in off-road scenarios depends heavily on semantic reasoning and fine-grained alignment between perception, language, and motion, making it a particularly challenging and compelling domain for language-grounded planning.

However, we observe that naively attaching natural language descriptions to existing off-road data provides limited benefit for trajectory planning. This weakness is evident in the top row of Fig.~\ref{fig:sft_refine}, where a model trained with original annotations generates generic scene descriptions (e.g., focusing on "dirt road" and "clear weather") that fail to capture specific vehicle actions or terrain geometry, leading to inconsistent trajectories. This weak alignment between language and action limits the model's ability to learn fine-grained motion behaviors, a limitation also noted in recent studies like Alpamayo~\cite{wang2025alpamayo}, which highlighted the necessity of action-aligned language supervision for effective planning.

In this work, we argue that the effectiveness of language in off-road driving critically depends on how language supervision is constructed and aligned with motion. Rather than relying on coarse or generic descriptions, we propose a data refinement framework that systematically restructures off-road driving data into language--action pairs explicitly grounded in vehicle trajectories and terrain geometry. Building upon this refined dataset (see Fig.~\ref{fig:sft_refine}, bottom row), we introduce a language-model-based trajectory generation approach, where the model reasons over visual observations and structured language representations to directly predict feasible off-road trajectories.

Furthermore, we observe that conventional planning metrics, primarily designed for urban driving, are insufficient for evaluating off-road performance. In off-road environments, strict point-wise agreement with a single ground-truth trajectory is often neither necessary nor desirable, as multiple trajectories may be equally valid depending on terrain structure. Instead, effective evaluation should capture whether predicted trajectories follow plausible traversable regions and reflect consistent terrain geometry. In particular, off-road driving critically depends on correctly modeling terrain elevation trends, such as uphill and downhill motions, which are largely ignored by conventional planar metrics. Therefore, we propose off-road-specific evaluation metrics and a training strategy that emphasize terrain awareness, drivable-region alignment, and consistency with terrain elevation profiles, in addition to traditional spatial accuracy.

In summary, our contributions are threefold:
\begin{itemize}
    \item We identify limitations in existing language-annotated off-road datasets and propose an efficient refinement strategy that aligns language supervision with vehicle motion and terrain geometry.
    \item We propose an off-road-specific, language-grounded training framework that generates 3D trajectories by jointly reasoning over visual observations, terrain geometry, and vehicle actions.
    \item We show that conventional evaluation protocols used in urban driving are inadequate for off-road scenarios and introduce terrain-aware metrics that assess traversability and elevation consistency for more reliable off-road planning evaluation.
\end{itemize}

%% file: sections/2_related_work.tex
\subsection{Off-Road Path Planning}
Off-road path planning presents significant challenges due to unstructured terrain and the absence of explicit road structures~\cite{Wijayathunga_Rassau_Chai_2023}. Traditional approaches primarily rely on geometric and kinematic constraints~\cite{Hu_Hu_Lu_Gong_Chen_2021,Hua_Niu_Yu_Zheng_Bai_Zhang_2022,Overbye_Saripalli_2021,Linker_Blass_2008}. Search-based methods, such as Hybrid A* and its variants, incorporate vehicle dynamics and terrain data for feasibility~\cite{Dolgov_2008}. Simultaneously, sampling-based algorithms like RRT*, T-RRT, and PRM have proven effective in addressing risk-awareness and computational efficiency in uncertain environments~\cite{Karaman_Frazzoli_2011,Devaurs_Siméon_Cortés_2016,Jeon_Karaman_Frazzoli_2011,Yin_Hu_Mourelatos_Gorsich_Singh_Tau_2023}.

To further enhance adaptability and multi-objective optimization, bio-inspired algorithms (GA, ACO) and learning-based techniques have been explored~\cite{Huang_Yuan_Liu_Tan_Wu_Wang_2023,Xiaodong_Tianqing_Kaixuan_Liyang_Jie_2022,Yépez-Ponce_Montalvo_2025}. Approaches utilizing Deep Reinforcement Learning (DRL) and Imitation Learning demonstrate improved performance in complex scenarios by balancing traversability and smoothness~\cite{Zhu_Li_Sun_Zhao_Xu_2019,Huang_Yuan_Liu_Tan_Wu_Wang_2023}.

Despite these advancements, conventional methods often rely on hand-crafted cost functions, explicit maps, or limited generalization capabilities~\cite{Vishwanath_Sujit_Saripalli_2022}. They frequently struggle with severe visual ambiguity and unseen terrain variations~\cite{Fortin_Gamache_Fecteau_Daum_Larriv’ee-Hardy_Pomerleau_Giguère_2024}, motivating the shift toward end-to-end paradigms that leverage richer semantic understanding.

\subsection{End-to-End Autonomous Driving}
End-to-end autonomous driving has achieved substantial progress by unifying perception, prediction, and planning within a single learning-based architecture~\cite{Chen_Wu_Chitta_Jaeger_Geiger_Li_2023}. Representative works, such as UniAD~\cite{Hu_Yang_Chen}, adopt a planning-oriented design that jointly reasons over visual features and motion forecasting to produce driving trajectories. While these approaches demonstrate strong performance in structured environments with clear lane topologies, they often struggle with generalization in open-world scenarios~\cite{Liao_Chen_Yin,Hallgarten_Zapata_Stoll_Renz_Zell_2024}. Purely data-driven imitation learning methods tend to learn statistical correlations rather than causal reasoning~\cite{Haan_Jayaraman_Levine_2019}, leading to failures in unstructured scenes involving long-horizon decision-making and rare, ambiguous events~\cite{Ochoa_Oh_Kwon_Domae_Matsubara_2024}.

To bridge this gap, recent research increasingly integrates Large Language Models (LLMs) and Vision-Language Models (VLMs) to endow driving systems with high-level semantic reasoning. Early efforts utilized VLMs primarily as auxiliary components or supervision sources~\cite{Narayanan_Maniar_Kalyanasundaram_Gandhi_Bhowmick_Krishna_2019}. However, loosely coupled architectures often face latency issues and information loss between the reasoning and control modules.

This limitation has led to the emergence of Vision-Language-Action (VLA) models, which unify perception, reasoning, and low-level control into a single transformer backbone. Inspired by robotics foundation models like RT-2~\cite{zitkovich2023rt}, VLA approaches in driving discretize continuous trajectories into action tokens. This allows the model to treat motion planning as an autoregressive generation task, effectively leveraging the world knowledge and logical capabilities of pre-trained LLMs. Recent agentic methods, such as LLaDA~\cite{li2024driving} and LaMPilot~\cite{ma2024lampilot}, further extend this paradigm by emphasizing closed-loop reasoning, instruction following, and adaptive policy adjustments in response to dynamic environments.

Building on this VLA-inspired philosophy, our work employs language-model-style generation for motion outputs. By integrating refined, action-aligned language supervision, we aim to enhance the system's ability to navigate complex, unstructured off-road terrain where geometric priors are insufficient.

%% file: sections/3_method.tex
\begin{figure}[t]
    \centering
    \includegraphics[width=\linewidth]{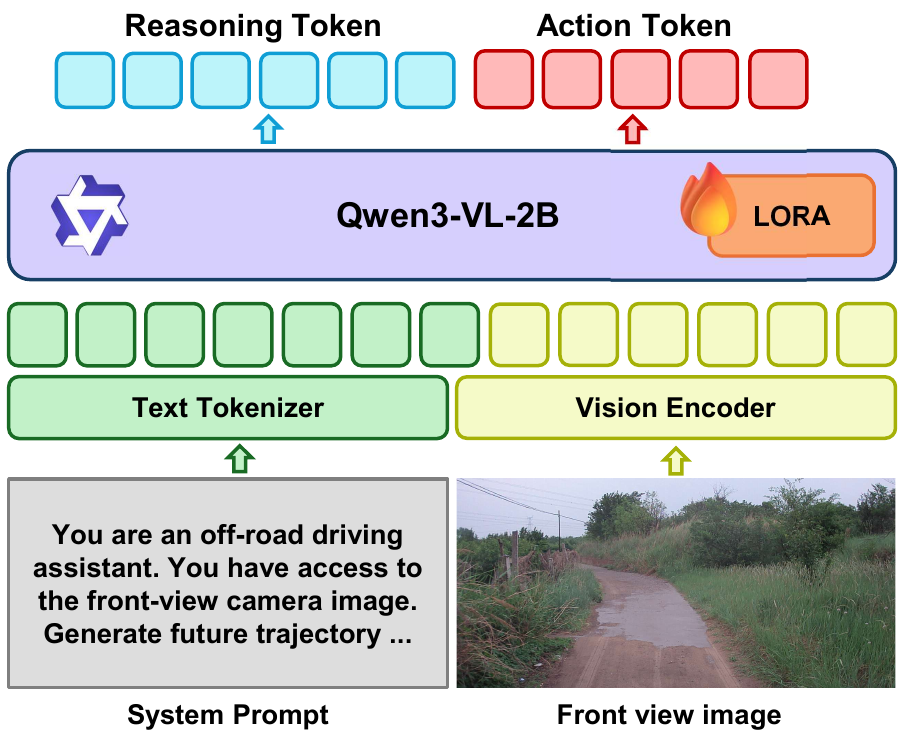}
    \caption{Overview of our proposed architecture. The model leverages a pre-trained Vision-Language Model (VLM) backbone with a trainable LoRA adapter to efficiently learn off-road driving capabilities.}
    \label{fig:model_architecture}
\end{figure}

To bridge the gap between high-level visual understanding and low-level geometric control in off-road environments, we propose an end-to-end vision-language architecture. As illustrated in Fig.~\ref{fig:model_architecture}, our framework builds upon a pre-trained Multimodal Large Language Model (MLLM) backbone. To efficiently adapt the model to the domain of off-road driving while preserving its generalization capabilities, we employed Low-Rank Adaptation (LoRA) on the linear layers of the decoder for fine-tuning~\cite{hu2022lora}.

The model processes a single front-view image along with a system instruction. It then operates in a unified autoregressive manner, generating a structured output sequence: first, a \emph{trajectory-aware scene description} that reasons about the terrain traversability and geometry, followed explicitly by \emph{future trajectory tokens} representing 3D waypoints. This ``reasoning-before-acting'' design encourages the model to ground its motion predictions in the linguistic understanding of the scene.

\subsection{Language Annotation Refinement}
\label{sec:language_refine}

\begin{figure}[t]
    \centering
    \includegraphics[width=\linewidth]{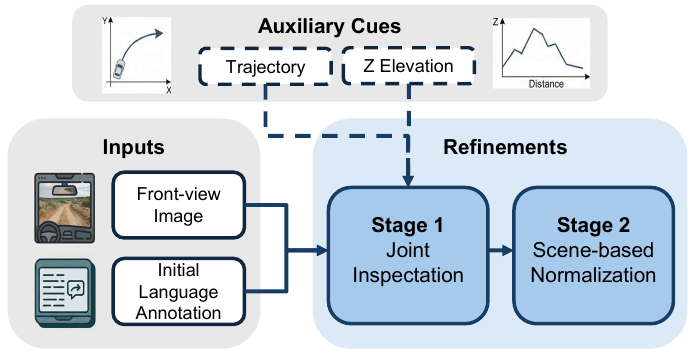}
    \caption{Overview of the Language Annotation Refinement Process. Initial
    scene-level annotations are refined using auxiliary cues,
    including future XY trajectories and Z-axis elevation profiles. The
    refinement consists of two stages: (1) joint inspection of visual, linguistic,
    and geometric information, and (2) scene-based normalization that aligns
    directional and elevation-related expressions through comparison with
    similar driving scenes.}
    \label{fig:language_refine}
\end{figure}

Existing off-road language annotations are typically scene-descriptive, focusing
on visual appearance rather than being explicitly aligned with vehicle actions
or terrain geometry. As a result, naively attaching such language to off-road
driving data provides weak supervision for trajectory learning. To address this
limitation, we propose a structured language annotation refinement pipeline that
transforms scene-level descriptions into terrain-aligned language,
while preserving a single front-view image input at inference time.

Inspired by the CoC dataset construction in Alpamayo-R1~\cite{wang2025alpamayo}, which highlights the
importance of capturing both driving decisions and environmental context, our
refinement process integrates visual observations with geometric ground truth at
annotation time. As illustrated in Fig.~\ref{fig:language_refine}, the pipeline
consists of two explicit stages: joint inspection and scene-based normalization.
Following prior works~\cite{xu2024drivegpt4, arai2025covla} that leverage large language models to generate or refine
supervision for language-based datasets, our refined
annotations are generated using Gemini-3-Flash-Preview, followed by human
verification to ensure correctness and consistency.

\vspace{2mm}
\noindent\textbf{Stage 1: Joint Inspection.}\hspace{2mm}
For each frame, the refinement process begins with a joint inspection of the
front-view image and the initial language annotation together with auxiliary
visualizations derived from ground-truth trajectories. These auxiliary cues,
shown in the top panel of Fig.~\ref{fig:language_refine}, include an XY-plane
trajectory plot that reveals the relative motion direction, and a terrain
elevation profile along the future trajectory that exposes vertical trends such
as uphill or downhill motion. Importantly, these auxiliary cues are used only
during annotation refinement and
are not provided to the model during training or inference.

\vspace{2mm}
\noindent\textbf{Stage 2: Scene-based Normalization.}\hspace{2mm}
Beyond generic language rewriting or data augmentation, our refinement process
explicitly addresses the fact that driving-related concepts are inherently
\emph{scene-relative}. As illustrated in the ``Scene-based Normalization'' module
of Fig.~\ref{fig:language_refine}, directional and elevation-related expressions
must be interpreted relative to local scene geometry and motion context. For
example, what constitutes a ``sharp turn'' or a ``steep slope'' varies
significantly across different off-road terrains.

To normalize such scene-dependent language, Gemini-3-Flash-Preview generates
candidate refined descriptions by comparing the current trajectory against
retrieved auxiliary visualizations from other scenes with similar horizontal motion patterns (e.g., comparable curvature and turning severity) and elevation
profiles. This retrieval-based comparison enables relative interpretation of
motion and elevation cues, rather than absolute or scene-agnostic descriptions.
This scene-based normalization mechanism distinguishes our approach from prior
language-guided refinement techniques and is critical for aligning language with
trajectory geometry in unstructured off-road settings.

\subsection{Language-Guided Trajectory Generation}
\label{sec:traj_gen}

Using the refined language annotations described in Sec.~\ref{sec:language_refine}, we train a
vision-language model to generate future off-road trajectories conditioned on a single front-view
camera image. The training is conducted in a supervised manner, where ground-truth language
annotations and corresponding ego trajectories are available for each sample.

Given an observation $o_t$ at time $t$, the model autoregressively generates an output sequence
consisting of two components: a refined scene description followed by an explicit ego trajectory.
The target sequence is defined as
\begin{equation}
y_t = \bigl[ w_1, \ldots, w_K, \langle \texttt{trajectory} \rangle,
(x_1,y_1,z_1), \ldots, (x_N,y_N,z_N) \bigr],
\label{eq:traj_output}
\end{equation}
where $\{w_i\}$ denote language tokens encoding scene context, terrain properties, and driving
direction, and $\{(x_i,y_i,z_i)\}_{i=1}^{N}$ represent the future ego trajectory as a sequence of
3D waypoints in the ego-centric coordinate frame.

We represent the trajectory as a sequence of action tokens, which are
treated identically to language tokens during autoregressive generation,
similar to prior embodied approaches such as RT-2~\cite{zitkovich2023rt}. As a result, the generated tokens directly represent the planned vehicle motion, capturing both horizontal direction and vertical elevation change. By generating the
trajectory conditioned on refined language, the model is encouraged to ground motion prediction
in scene-relative semantic cues.

Model training follows a standard supervised fine-tuning paradigm based on autoregressive
maximum likelihood estimation. Let $\theta$ denote the model parameters. The optimization
objective is
\begin{equation}
\mathcal{L}(\theta)
= - \sum_{i=1}^{|y_t|} \log p_\theta \bigl( y_{t,i} \mid o_t, y_{t,<i} \bigr),
\label{eq:sft_loss}
\end{equation}
where $y_{t,i}$ denotes the $i$-th token in the output sequence. This objective jointly supervises
language generation and trajectory planning, enabling the model to learn off-road motion
generation directly from refined, action-aligned language annotations.

\subsection{Preference Optimization with Geometry-Aware Hard Negatives}
\label{sec:orpo}

\begin{figure*}[t]
    \centering
    \includegraphics[width=\linewidth]{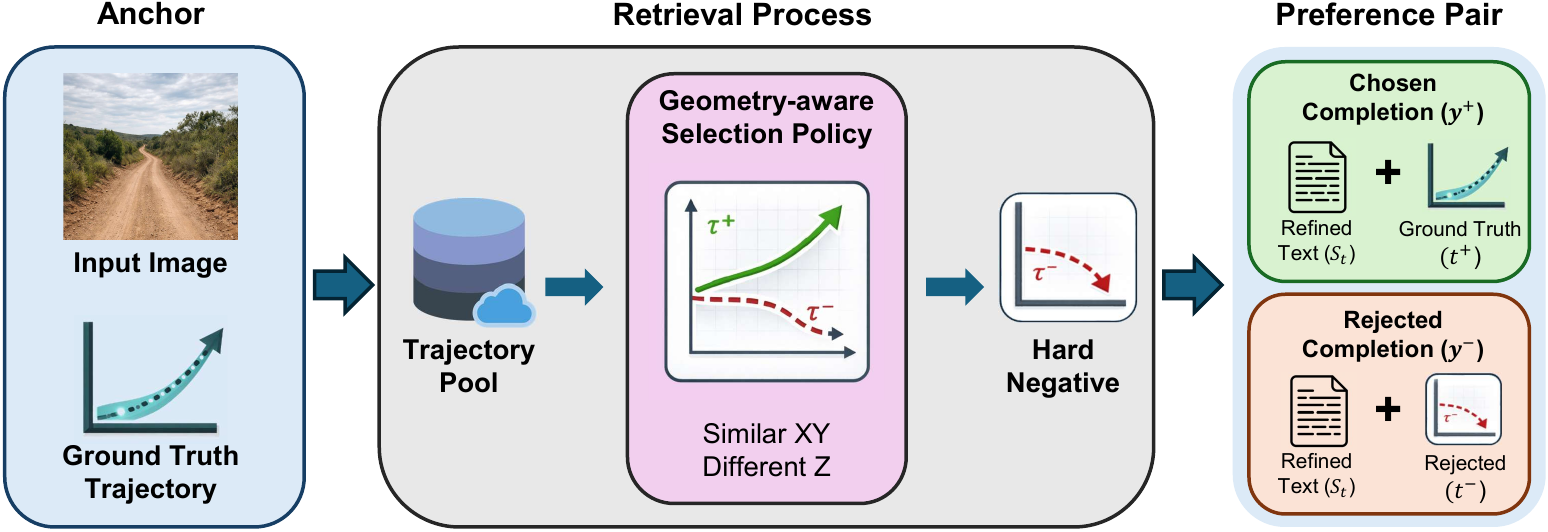}
    \caption{Illustration of \textbf{geometry-aware hard negative mining} for
    preference optimization. To create preference pairs for ORPO, we retrieve a
    ``rejected'' trajectory ($\tau^-$) from the \textbf{same scene}. We employ a
    discrepancy-aware scoring function that maximizes the elevation difference
    ($\Delta_z$) while minimizing the planar deviation ($\Delta_{xy}$) with
    respect to the ground-truth trajectory ($\tau^+$). This produces hard
    negatives that are spatially deceptive in the XY plane but exhibit
    contradictory vertical behaviors.}
    \label{fig:orpo_process}
\end{figure*}

While the model trained in Sec.~\ref{sec:traj_gen} learns to generate trajectories
from refined language--trajectory pairs via supervised learning, off-road
planning demands stronger alignment with terrain feasibility, particularly along
the vertical axis. Elevation changes such as uphill, downhill, and uneven ground
directly affect drivability, and these factors are explicitly encoded as priors
in our refined language annotations. To further transfer these priors into
trajectory generation, we employ Odds Ratio Preference Optimization (ORPO)~\cite{hong2024orpo} with a
geometry-aware hard negative mining strategy.

As illustrated in Fig.~\ref{fig:orpo_process}, we construct preference data by
pairing each prompt with a \emph{chosen} completion and a \emph{rejected}
completion. The prompt consists of the same single front-view camera image and
instruction used during supervised training. For a frame at time $t$, the chosen
completion (green box in Fig.~\ref{fig:orpo_process}) is formed by concatenating
the refined scene description $S_t$ with the ground-truth trajectory
$\tau_t^{+}$. The rejected completion (red box in
Fig.~\ref{fig:orpo_process}) keeps the same $S_t$ but replaces the trajectory with
a mismatched trajectory $\tau_t^{-}$ retrieved from the \textbf{same off-road
scene}:

\begin{equation}
y_t^{+} = \bigl[S_t \;\Vert\; \langle \texttt{trajectory} \rangle \;\Vert\; \tau_t^{+}\bigr],
\label{eq:orpo_positive}
\end{equation}
\begin{equation}
y_t^{-} = \bigl[S_t \;\Vert\; \langle \texttt{trajectory} \rangle \;\Vert\; \tau_t^{-}\bigr].
\label{eq:orpo_negative}
\end{equation}
By sampling negatives from the same scene, the preference signal focuses entirely on fine-grained trajectory quality within a consistent visual context, rather than obvious background mismatches.

To construct informative hard negatives, we scan the scene-local trajectory pool, as shown in the central module of Fig.~\ref{fig:orpo_process}. We retain only frames with non-empty refined language and trajectories containing at least two valid 3D waypoints. Crucially, we aim to find negatives that are ``deceptive'': they should follow a similar path to the ground truth in the horizontal (XY) plane but exhibit incorrect behavior in the vertical (Z) axis. To achieve this, we employ a \textbf{discrepancy-aware scoring function} $\mathcal{S}(\tau, \tau^+)$ that penalizes XY deviation while rewarding Z deviation:
\begin{equation}
\mathcal{S}(\tau, \tau^+) = \frac{1}{N}\sum_{i=1}^{N}|z_i - z_i^+| - \lambda_{\mathrm{geo}} \cdot \frac{1}{N}\sum_{i=1}^{N}\|\mathbf{p}_{xy,i} - \mathbf{p}_{xy,i}^+\|_2,
\label{eq:scoring_func}
\end{equation}
where $\mathbf{p}_{xy}$ denotes the 2D position vector, and $\lambda_{\mathrm{geo}}$ is a balancing coefficient. We select the candidate with the highest score as the rejected trajectory $\tau^-$. This strategy forces the model to attend to visual cues related to terrain geometry—such as texture gradients and vanishing point—rather than relying solely on road curvature.

Given a preference pair $(y_t^{+}, y_t^{-})$, we optimize the model using the ORPO~\cite{hong2024orpo} objective. Unlike standard preference alignment methods that require a reference model, ORPO incorporates an odds ratio penalty directly into the supervised loss. Let $\pi_\theta$ denote the policy parameterized by $\theta$. The odds of generating a completion $y$ given input $x$ are defined as
\begin{equation}
\mathrm{odds}_\theta(y \mid x) = \frac{\pi_\theta(y \mid x)}{1 - \pi_\theta(y \mid x)}.
\label{eq:odds_def}
\end{equation}
The final loss function is formulated as
\begin{equation}
\mathcal{L}_{\mathrm{ORPO}}(\theta)
= \mathcal{L}_{\mathrm{SFT}} + \lambda_{\mathrm{ORPO}} \mathcal{L}_{\mathrm{odds}},
\label{eq:orpo_loss}
\end{equation}
where $\mathcal{L}_{\mathrm{SFT}}$ is the negative log-likelihood of the chosen label, and $\mathcal{L}_{\mathrm{odds}}$ maximizes the log-odds ratio gap between the chosen and rejected completions:
\begin{equation}
\mathcal{L}_{\mathrm{odds}}
= - \log \sigma \bigl(
\log \mathrm{odds}_\theta(y_t^{+}\mid x_t)
-
\log \mathrm{odds}_\theta(y_t^{-}\mid x_t)
\bigr).
\label{eq:odds_loss}
\end{equation}
Here, $x_t$ denotes the shared prompt, $\sigma(\cdot)$ is the sigmoid function, and $\lambda_{\mathrm{ORPO}}$ is a hyperparameter weighing the preference signal. This objective explicitly penalizes trajectories that violate terrain-aware priors encoded in the refined language, while reinforcing trajectories consistent with both horizontal motion and terrain elevation trends.

\subsection{Off-Road Evaluation Metrics}
\label{sec:offroad_metrics}
In off-road driving scenarios, multiple trajectories may be feasible, and the
primary requirement is navigating through free-space while respecting terrain
geometry. We therefore evaluate predicted trajectories along two independent
axes: (i) a traversability-based metric measuring adherence to navigable regions,
and (ii) a vertical trend metric measuring consistency with terrain elevation.
Both metrics avoid direct geometric matching to a single ground-truth (GT)
trajectory.

\vspace{2mm}
\noindent\textbf{Traversability Compliance Score.}\hspace{2mm}
To assess how well the trajectory respects the drivable area, we compute the
clearance $c_i$ from each waypoint to the nearest non-traversable region based
directly on the ground-truth occupancy labels. To emphasize failures in critical
regions, we focus on the subset of lowest clearance values (e.g., top-$K$
closest encounters) and define a violation score as
\begin{equation}
\mathrm{Risk}
= \frac{1}{|\mathcal{C}|}\sum_{c \in \mathcal{C}} \exp(-\alpha c),
\label{eq:risk}
\end{equation}
where $\mathcal{C}$ denotes the selected set of lowest clearance values and
$\alpha$ controls the sensitivity to obstacle proximity. Based on this risk
measure, we compute a base compliance score:
\begin{equation}
S_{\mathrm{base}} = \mathrm{clip}(1 - \mathrm{Risk},\,0,\,1).
\label{eq:base_score}
\end{equation}
To strictly penalize trajectories that traverse into infeasible terrain, we
additionally compute a length-weighted violation ratio for a set of strict
clearance thresholds $\mathcal{D}=\{d_1,\dots,d_K\}$. Let $r_{\mathrm{bad}}(d)$
denote the fraction of the trajectory length violating threshold $d$, defined as
\begin{equation}
r_{\mathrm{bad}}(d) = \frac{\sum_{i:\,\mathrm{seg}_i\,\mathrm{bad}} \|\mathbf{p}_{i+1} - \mathbf{p}_{i}\|_{xy}}{\sum_{i} \|\mathbf{p}_{i+1} - \mathbf{p}_{i}\|_{xy}},
\label{eq:r_bad}
\end{equation}
where a segment $i$ between consecutive waypoints $\mathbf{p}_i$ and $\mathbf{p}_{i+1}$
is marked bad if either endpoint has clearance below $d$ or lies on a non-traversable
region. We define a compliance penalty as
\begin{equation}
P = \max\left(0,\; 1 - \lambda \cdot \frac{1}{|\mathcal{D}|}
\sum_{d \in \mathcal{D}} r_{\mathrm{bad}}(d)\right),
\label{eq:penalty}
\end{equation}
where $\lambda$ determines the penalty severity. The final traversability
compliance score is given by
\begin{equation}
S_{\mathrm{trav}} = S_{\mathrm{base}} \cdot P,
\label{eq:trav_score}
\end{equation}
where higher values indicate better adherence to traversable regions.

\vspace{2mm}
\noindent\textbf{Elevation Consistency Metric.}\hspace{2mm}
To evaluate elevation reasoning, we compare the predicted and ground-truth (GT)
height profiles along the overlapped traveled distance parameterized by the XY
arc-length $s$. After optional mean-centering, we compute the following error
terms:
\begin{equation}
\mathrm{MSE}_z, \quad \mathrm{RMSE}_z, \quad \mathrm{MSE}_{dz},
\label{eq:z_errors}
\end{equation}
where $\mathrm{MSE}_{dz}$ measures the discrepancy between elevation slopes
$\frac{dz}{ds}$ and captures differences in local ascent and descent behavior.
Since these error terms have different numerical scales, we introduce weighting
coefficients to balance their contributions when aggregating them into a single
score. The final elevation consistency metric (lower is better) is defined as
\begin{equation}
S_z = w_1 \, \mathrm{MSE}_z
    + w_2 \, \mathrm{RMSE}_z
    + w_3 \, \mathrm{MSE}_{dz},
\label{eq:z_score}
\end{equation}
where $w_1$, $w_2$, and $w_3$ are positive scalar weights that normalize
the relative scales of each term. These weights are fixed across all experiments
to ensure fair comparison between different models.

Together, the traversability compliance score $S_{\mathrm{trav}}$ (higher is
better) and the elevation consistency score $S_z$ (lower is better) assess
whether a predicted trajectory remains within feasible terrain boundaries and
faithfully follows the underlying ground geometry.

%% file: sections/4_experiments.tex
We evaluate the proposed language refinement and preference optimization
strategies on the ORAD-3D dataset~\cite{ORAD-3D}. Our experiments focus on verifying the
importance of trajectory-aligned language supervision and assessing off-road
planning performance under both standard planar metrics and terrain-aware
evaluation metrics.

\subsection{Datasets and Evaluation Metrics}

\noindent\textbf{Dataset.}\hspace{2mm}
All experiments are conducted on the ORAD-3D dataset~\cite{ORAD-3D}.
Unlike most existing off-road datasets designed primarily for perception or SLAM tasks, to the best of our knowledge, ORAD-3D is the only publicly available benchmark that jointly provides synchronized front-view images and ego-centric 3D trajectories paired with natural language annotations, which are essential for enabling semantic reasoning in unstructured environments.
We follow the official split protocol, using 100 scenes for training (39,489 images), 15 scenes for validation (5,717 images), and 29 scenes for testing (12,164 images).

\vspace{2mm}
\noindent\textbf{Evaluation Metrics.}\hspace{2mm}
We evaluate trajectory planning using standard planar metrics—L2 displacement errors at 1s, 2s, and 3s horizons, along with average error and failure rate—which are widely adopted in benchmarks like nuScenes~\cite{nuscenes} for assessing horizontal trajectory quality. However, these metrics do not capture terrain feasibility or elevation consistency critical to off-road environments. Therefore, we primarily rely on our proposed terrain-aware off-road metrics: \emph{Traversability Compliance} and \emph{Elevation Consistency}. Since prior off-road works have not explicitly addressed the challenge of generating \textit{XYZ} trajectories from a single front-view image, we compare our approach against recent state-of-the-art open vision-language models adapted to this task.

\subsection{Implementation Details}

\noindent\textbf{Training Setup.}\hspace{2mm}
Our fine-tuning base model is \textbf{Qwen3-VL-2B}~\cite{yang2025qwen3}.
We apply Low-Rank Adaptation (LoRA) adapters with a rank of 32 to the linear layers of the decoder.
Fine-tuning is performed on four NVIDIA H100 GPUs for 5 epochs, with a global batch size of 32 and a learning rate of $1.0 \times 10^{-4}$.

\vspace{2mm}
\noindent\textbf{Trajectory Generation for Baselines and Our Model.}\hspace{2mm}
To maximize the performance of open models and enable a fair comparison, we
adopt an in-context demonstration strategy for all VLM baselines. Specifically,
we provide three image--trajectory pairs sampled from the training set as
in-context examples, which explicitly specify the expected trajectory format and
generation pattern. While providing these pairs, we automatically selected left, right and straight trajectory pairs so that the models can understand how to generate future trajectory. Test-time trajectories are then generated under the same
prompt. This design encourages baselines to produce well-formed trajectories,
rather than being limited by naive or underspecified prompting.

In contrast, our model is explicitly fine-tuned for trajectory generation using
refined language--trajectory pairs. As a result, it does not require any
in-context examples during inference. The model takes only a single front-view
image as input and directly outputs reasoning and the predicted trajectory,
reflecting its stronger alignment with the off-road trajectory planning task.

We set the off-road metric parameters to $\alpha=3$ and $K=20$ for traversability
risk, and keep the elevation consistency weights fixed at
$(w_1,w_2,w_3)=(5,1,10)$ across all experiments.

\subsection{Results}

\begin{figure*}[tb]
  \centering
  \includegraphics[width=\linewidth]{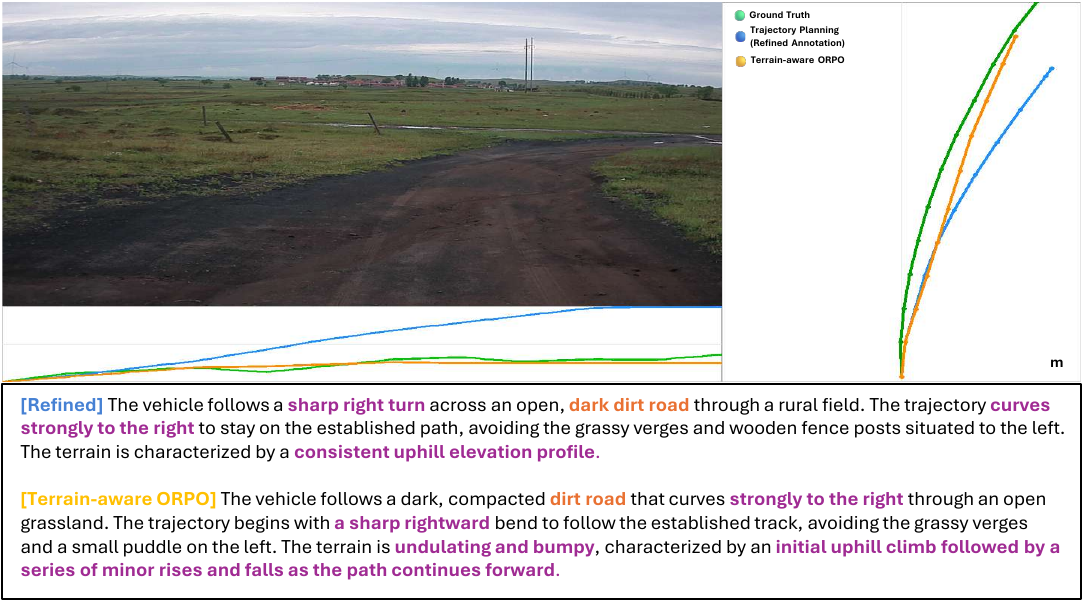}
  \caption{Qualitative comparison between Refined SFT and Terrain-aware ORPO.
  Top-left shows the input front-view image. The top-right panels visualize the
  predicted trajectories in the XY plane, while the bottom panel shows the
  corresponding elevation profiles plotted as relative height changes
  ($z - z_0$) along the traveled distance. Compared to Refined SFT, Terrain-aware
  ORPO produces trajectories that not only remain within traversable regions
  but also better capture elevation trends, such as uphill motion and local
  undulations, consistent with the underlying terrain geometry.}
  \label{fig:orpo}
\end{figure*}

\begin{table}[t]
\centering
\caption{VLM-based path planning results on ORAD-3D under standard planar metrics.}
\label{tab:refine_l2}
\small
\setlength{\tabcolsep}{4pt}
\resizebox{\columnwidth}{!}{%
\begin{tabular}{lccccc}
\toprule
\textbf{Method} &
\multicolumn{4}{c}{\textbf{L2 (m)$\downarrow$}} &
\textbf{Failure} \\
 & \textbf{1s} & \textbf{2s} & \textbf{3s} & \textbf{Avg.} & \textbf{Rate (\%)$\downarrow$} \\
\midrule
Qwen2.5-VL   & 1.95 & 2.21 & 2.90 & 2.35 & 82.55 \\
OpenEMMA~\cite{xing2025openemma}     & 1.73 & 1.92 & 2.75 & 2.13 & 57.56 \\
LightEMMA~\cite{lightemma}    & 1.72 & 1.88 & 2.63 & 2.08 & 55.63 \\
\midrule
SFT (raw lang.)      & 0.92 & 0.95 & 1.15 & 1.01 & 6.67 \\
\rowcolor{cyan!15}
Refined SFT (ours)  & 0.88 & 0.91 & 1.13 & 0.97 & 6.05 \\
\bottomrule
\end{tabular}
}
\end{table}

\begin{table}[t]
\centering
\caption{Comparison under the proposed off-road metrics.
Higher is better for Traversability Compliance; lower is better for Elevation
Consistency.}
\label{tab:offroad_metric}
\small
\setlength{\tabcolsep}{4pt}
\resizebox{\columnwidth}{!}{%
\begin{tabular}{lcc}
\toprule
\textbf{Model} &
\textbf{Traversability} &
\textbf{Elevation} \\
 & \textbf{Compliance} $\uparrow$ & \textbf{Consistency} $\downarrow$ \\
\midrule
\rowcolor{cyan!15}
Terrain-aware ORPO (ours) & 0.644 & 0.322 \\
Refined SFT (ours) & 0.621 & 0.428 \\
\midrule
Gemini-3-Flash-Preview & 0.589 & 0.639 \\
GPT-4o                & 0.538 & 0.559 \\
LLaMA-4-Scout         & 0.253 & 0.707 \\
Qwen3-VL-8B-Instruct  & 0.452 & 0.647 \\
Qwen3-VL-32B-Instruct & 0.503 & 0.575 \\
Qwen3-VL-2B-Instruct  & 0.152 & 0.676 \\
\bottomrule
\end{tabular}
}
\end{table}

We present experimental results to analyze (i) the effect of trajectory-aligned
language refinement on off-road trajectory planning, and (ii) the benefit of
terrain-aware preference optimization for improving both planar feasibility and
vertical consistency.

\vspace{2mm}
\noindent\textbf{Effect of Language Refinement.}\hspace{2mm}
We first evaluate the impact of language refinement by comparing supervised
fine-tuning with raw language annotations (SFT) against fine-tuning with refined
language annotations (Refined SFT), using identical model architecture and
training settings.

Quantitative results under standard planar metrics are reported in
Table~\ref{tab:refine_l2}. Refined SFT substantially outperforms both raw-language
SFT and recent VLM-based baselines across all horizons, reducing average L2
error and failure rate by a large margin. These improvements indicate that
refined annotations provide more effective supervision for trajectory
generation, even when evaluated using conventional on-road–style metrics.

Figure~\ref{fig:sft_refine} provides qualitative evidence supporting this trend.
With raw annotations, the generated trajectories often lack clear directional
intent and exhibit inconsistent curvature, reflecting the weak alignment
between generic scene descriptions and vehicle motion. In contrast, refined
annotations explicitly encode action-relevant cues such as turning direction
and obstacle avoidance. As a result, the model generates trajectories that
follow coherent motion patterns and more closely match the ground-truth
geometry. Notably, the refined language descriptions themselves also become more
directional and action-oriented, demonstrating that the refinement process
improves both trajectory planning and language–motion alignment.

Together, these results confirm that naive language supervision is insufficient
for off-road trajectory learning, and that action- and terrain-aligned language
refinement plays a critical role in improving planning performance.

\vspace{2mm}
\noindent\textbf{Effect of Terrain-Aware Optimization.}\hspace{2mm}
While refined language supervision significantly improves planar trajectory
accuracy, off-road driving further requires consistency with terrain geometry,
particularly along the vertical axis. To assess this aspect, we evaluate models
using the proposed off-road metrics: Traversability Compliance and Elevation
Consistency.

As shown in Table~\ref{tab:offroad_metric}, Refined SFT already improves
traversability compared to open-source VLM baselines, indicating that
trajectory-aligned language helps the model stay within feasible regions.
However, Terrain-aware ORPO achieves additional gains, yielding the highest
Traversability Compliance and the lowest Elevation Consistency error among all
evaluated methods.

Figure~\ref{fig:orpo} qualitatively illustrates these improvements. When trained
with the same refined language data, Terrain-aware ORPO produces trajectories
that more faithfully follow both the horizontal path structure and the vertical
terrain profile. In particular, the elevation plots shown as relative height
changes ($z - z_0$) reveal that Terrain-aware ORPO captures coherent uphill and downhill trends,
as well as local undulations, whereas Refined SFT often exhibits smoothed or
inconsistent elevation behavior. These differences are also reflected in the
generated language descriptions, where Terrain-aware ORPO produces more detailed
and accurate accounts of elevation changes, such as initial climbs followed by
minor rises and falls.

Finally, comparisons with recent open-source vision–language models under the
proposed off-road metrics further highlight the advantage of our approach.
Despite their strong general reasoning capabilities, these models struggle to
produce trajectories that are both traversable and elevation-consistent. In
contrast, Terrain-aware ORPO consistently achieves superior performance,
demonstrating that preference optimization with terrain-aware hard negatives
effectively transfers elevation priors into trajectory generation.

Overall, these results show that refined language supervision establishes a
strong foundation for off-road trajectory planning, while terrain-aware
preference optimization further enhances robustness by explicitly enforcing
consistency with terrain geometry. Together, they enable more reliable and
terrain-aware off-road planning than existing VLM-based approaches.

%% file: sections/5_conclusion.tex
This work highlights the critical role of aligning language supervision with terrain geometry for effective off-road trajectory planning . We proposed a language refinement framework and a retrieval-based preference optimization strategy to overcome the limitations of existing datasets, which often lack spatial precision. Furthermore, we introduced new evaluation metrics specifically designed to assess off-road traversability and safety. Our approach demonstrates superior performance on the ORAD-3D benchmark compared to state-of-the-art VLMs, particularly in challenging terrain conditions. These results validate the importance of our terrain-aware alignment and evaluation protocols. Moving forward, we plan to incorporate multi-view temporal inputs and richer terrain representations to enable fully closed-loop off-road autonomy.